\documentclass[conference]{IEEEtran}
\IEEEoverridecommandlockouts

\usepackage{cite}
\usepackage{amsmath,amssymb,amsfonts}
\usepackage{algorithmic}
\usepackage{graphicx}
\usepackage{textcomp}
\usepackage{xcolor}
\def\BibTeX{{\rm B\kern-.05em{\sc i\kern-.025em b}\kern-.08em
    T\kern-.1667em\lower.7ex\hbox{E}\kern-.125emX}}
\usepackage{url}
\usepackage{subcaption}
\usepackage[hidelinks]{hyperref}

\begin{document}

\title{UMETTS: A Unified Framework for Emotional Text-to-Speech Synthesis with Multimodal Prompts
\thanks{\textsuperscript{$\star$}Equal contribution (alphabetical). \textsuperscript{$\dagger$}Corresponding author: pengxiaojiang@sztu.edu.cn.
Funding: NSFC China (62176165), Shenzhen Higher Education Support (20220718110918001), SZTU Top Talent (GDRC202131). 
}
}
\author{
Zhi-Qi Cheng\textsuperscript{1, $\star$},
\IEEEauthorblockN{Xiang Li\textsuperscript{2, $\star$}, Jun-Yan He\textsuperscript{3}, Junyao Chen\textsuperscript{2}, Xiaomao Fan\textsuperscript{2},  \\ Xiaojiang Peng\textsuperscript{2, $\dagger$},\IEEEmembership{Senior Member,~IEEE},  Alexander G. Hauptmann\textsuperscript{4}}
\IEEEauthorblockA{
\textit{\textsuperscript{1}University of Washington},
\textsuperscript{2}\textit{Shenzhen Technology University},
\textit{\textsuperscript{3}DAMO Academy, Alibaba Group},
\textit{\textsuperscript{4}Carnegie Mellon University}
}
}

\maketitle

\begin{abstract}
Emotional Text-to-Speech (E-TTS) synthesis has garnered significant attention in recent years due to its potential to revolutionize human-computer interaction. However, current E-TTS approaches often struggle to capture the intricacies of human emotions, primarily relying on oversimplified emotional labels or single-modality input. In this paper, we introduce the Unified Multimodal Prompt-Induced Emotional Text-to-Speech System (UMETTS), a novel framework that leverages emotional cues from multiple modalities to generate highly expressive and emotionally resonant speech. The core of UMETTS consists of two key components: the Emotion Prompt Alignment Module (EP-Align) and the Emotion Embedding-Induced TTS Module(EMI-TTS). ~(1)~EP-Align employs contrastive learning to align emotional features across text, audio, and visual modalities, ensuring a coherent fusion of multimodal information.~(2)~Subsequently, EMI-TTS integrates the aligned emotional embeddings with state-of-the-art TTS models to synthesize speech that accurately reflects the intended emotions. 
Extensive evaluations show that UMETTS achieves significant improvements in emotion accuracy and speech naturalness, outperforming traditional E-TTS methods on both objective and subjective metrics. 
To facilitate reproducibility and further research, we have made our code publicly available at \url{https://github.com/KTTRCDL/UMETTS}.

\end{abstract}

\begin{IEEEkeywords}
Emotional Text-to-Speech, Multimodal Synthesis, Contrastive Learning, Expressive Speech Synthesis, Human-Computer Interaction.
\end{IEEEkeywords}

\section{Introduction} Emotional text-to-speech (E-TTS) technology has significantly enhanced human-computer interaction by adding emotional depth to synthetic voices, thereby creating more engaging and empathetic virtual agents \cite{tan2021survey, zhang2023iemotts, cui2021emovie}. This technology has the potential to revolutionize industries such as entertainment, education, healthcare, and customer service.

\begin{figure}
\centering
\includegraphics[width=\linewidth]{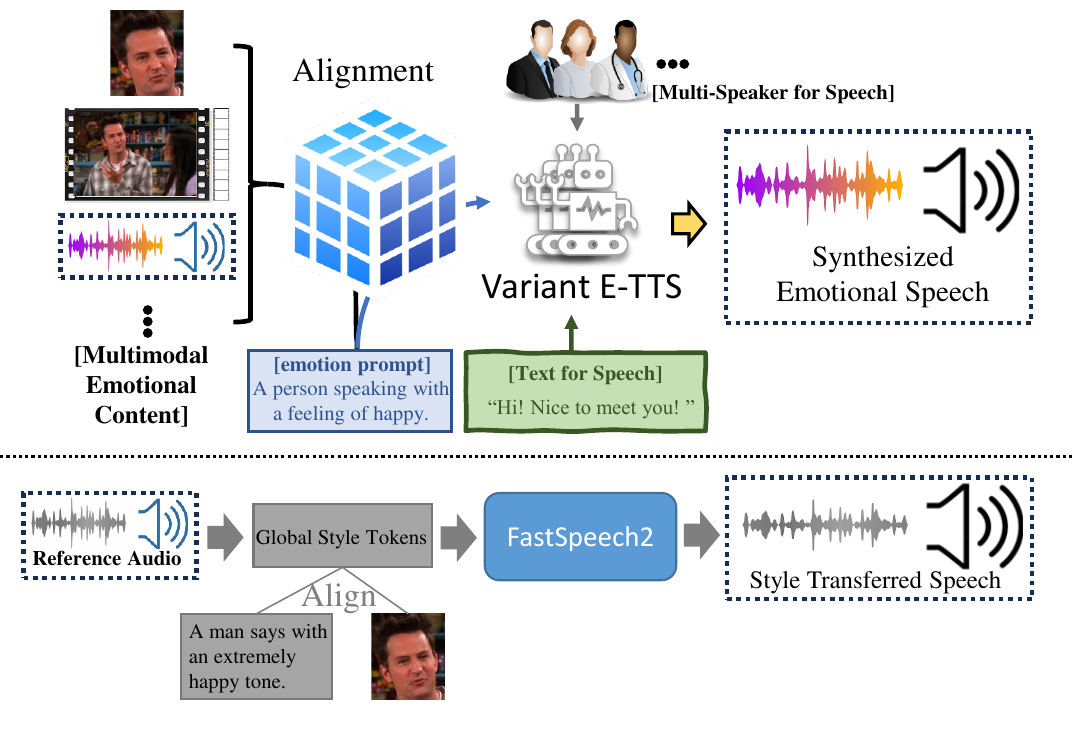}
\vspace{-8mm}
\caption{\small Top: The \textit{UMETTS} framework adeptly synthesizes speech by incorporating emotional cues from multiple modalities, ensuring the output speech consistently conveys the intended emotions. 
This capability is exemplified through our multimodal references, accessible by clicking the respective links: \href{https://github.com/KTTRCDL/UMETTS/tree/main/demo/static/demo_image/reference_image.png}{[Image]}, \href{https://github.com/KTTRCDL/UMETTS/tree/main/demo/static/demo_video/reference_video.mp4}{[Video]}, \href{https://github.com/KTTRCDL/UMETTS/tree/main/demo/static/demo_audio/reference_audio.wav}{[Audio]}, and \href{https://github.com/KTTRCDL/UMETTS/tree/main/demo/static/demo_audio/demo.wav}{[Synthesized Speech]}.
~[Click on brackets to access source files]. 
Bottom: Emotion speech synthesizes by  Style Transfer Model.}
\label{fig:intro}
\vspace{-7mm}
\end{figure}

Despite advancements in TTS technologies that produce naturalistic speech, the focus has primarily been on linguistic accuracy rather than capturing emotional nuances \cite{concatenationsynthesis, yoshimura1999simultaneous, sherstinsky2020fundamentals, wang2017tacotron, vaswani2017attention, shen2018tacotron2, ren2019fastspeech, ren2020fastspeech, dinh2014nice, vits, miao2021efficienttts}. Current E-TTS approaches often rely on emotion labels or reference speech, which can be limiting due to oversimplified representations and single-modality inputs \cite{emospeech, lee2017emotional, cui2021emovie, zhang2023iemotts, guo2023emodiff, Controllable_emotion_transfer, wu2021cross, IEEE_instructTTS, guo2023prompttts}. Furthermore, style transfer studies within TTS predominantly use Global Style Tokens (GST) extracted from reference audio, which may not fully disentangle speaker characteristics from emotional and prosodic elements, leading to suboptimal performance in out-of-domain scenarios \cite{generspeech, IEEE_metts}, as shown at the bottom of Figure\ref{fig:intro}.

\begin{figure*}
  \centering
  
  \includegraphics[width=\textwidth]{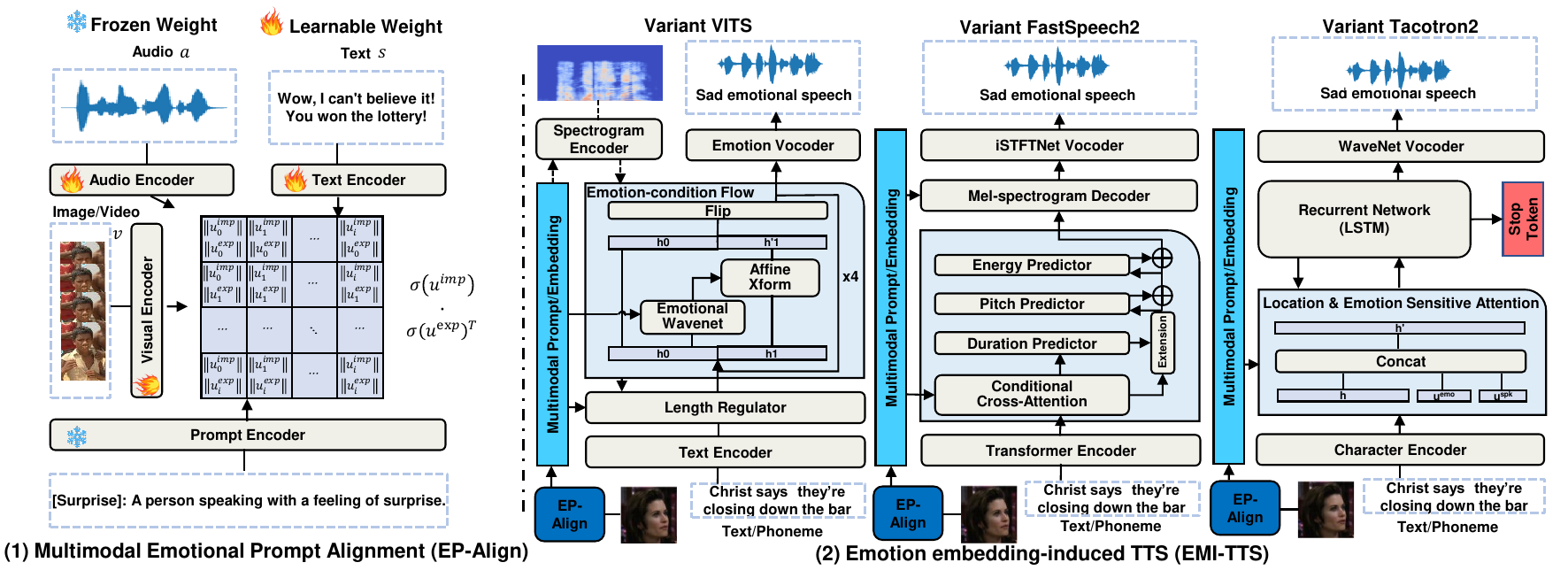}
  \vspace{-7mm}
  \caption{\small  The Overview of UMETTS Framework. UMETTS consists of two components: 1) Multimodal Emotional Prompt Alignment (EP-Align) and 2) Emotion embedding-induced TTS (EMI-TTS). EP-Align involves multimodal emotional presentation alignment, empowering EMI-TTS with multimodal emotional information-inducing audio synthesis.}
  \label{fig:model_overview}
  \vspace{-6mm}
\end{figure*}

To address these challenges, we present the Unified Multimodal, Prompt-Induced Emotional Text-to-Speech System (UMETTS), a groundbreaking framework designed to elevate the expressiveness of synthesized speech by incorporating multimodal cues encompassing text, audio, and visual information, as shown at the top of Figure \ref{fig:intro}. UMETTS comprises two critical components: the Emotion Prompt Alignment Module (EP-Align) and the Emotion Embedding-Induced TTS Module (EMI-TTS). EP-Align enables the seamless fusion of multimodal information by aligning emotional data across modalities through a cue anchoring mechanism. EMI-TTS harnesses these aligned emotional embeddings to generate expressive and emotionally resonant speech suitable for a wide range of applications. Extensive testing across diverse datasets demonstrates UMETTS's superior performance in both objective and subjective evaluations, positioning it as a leading solution in Emotional TTS technologies. Our key contributions are threefold: 
\begin{itemize} 
\item We introduce EP-Align, a contrastive learning-based module that synchronizes emotional features across modalities, addressing distribution discrepancies and enhancing emotional speech generation. 
\item We develop EMI-TTS, which integrates advanced TTS models with aligned emotional embeddings, resulting in speech that authentically reflects intended emotions and enhances user engagement. 
\item We validate UMETTS through extensive evaluations, confirming its ability to maintain emotional consistency and preserve speaker characteristics across various scenarios, showcasing its potential to revolutionize human-computer interactions. 
\end{itemize}

\section{Methodology}
The UMETTS framework aims to generate emotionally expressive speech by leveraging multimodal emotion cues from vision, audio, and text in a data-efficient manner. The system is composed of two core modules: the \textit{Emotion Prompt Alignment Module (EP-Align)} and the \textit{Emotion Embedding-induced TTS (EMI-TTS)}. Figure \ref{fig:model_overview} provides an overview of the architecture. In this section, we detail the components, their interactions, and the different variants of the TTS models supported within UMETTS.

\subsection{Emotion Prompt Alignment Module (EP-Align)}
EP-Align is responsible for extracting and aligning emotional representations from multiple modalities (vision, audio, text, and prompts). It processes the multimodal emotional data tuple \( Tup^{emo} = \langle v, a, s, p \rangle \), where \( v \) represents visual data, \( a \) is audio input, \( s \) denotes text input, and \( p \) is an emotional prompt label. Each modality is processed by modality-specific encoders, denoted by \( \mathcal{E} = \{\mathcal{E}^{vis}, \mathcal{E}^{audio}, \mathcal{E}^{tex}, \mathcal{E}^{prop} \} \), which extract emotion features. These features are aligned into a unified emotion embedding \( u^{emo} \), inspired by \cite{CLIP, cheng2024emotionllamamultimodalemotionrecognition, SZTUMER2024}, and subsequently passed to EMI-TTS for synthesis.

\subsubsection{Multimodal Feature Extraction}
Each encoder processes its input modality by transforming the raw input into a shared internal feature space. For each modality \( \mu \in \{vis, audio, tex, prop\} \), the extracted feature representations \( f^\mu \) are given by: $f^\mu = \mathcal{E}^\mu(x^\mu)$, where \( x^\mu \) represents the input data for modality \( \mu \). The shared embedding space is computed using a learnable projection matrix: $u^\mu = f^\mu \cdot W^{\mu-pro}.$

\subsubsection{Prompt Anchoring and Alignment}
To ensure consistency across modalities, a prompt-anchoring scheme is used to align the emotion features from different modalities (vision, audio, and text). The prompt embedding \( u^{prop} \) is selected based on the alignment modality \( \eta \) and is calculated as: $u^{prop} = f^{prop} \cdot W^{pro-\eta}, \eta \in \{vis, audio, tex\}.$

The alignment is further refined by computing the cosine similarity between the explicit emotion embeddings \( u^{exp} \) from the prompts and the implicit emotion embeddings \( u^{imp} \) from vision, audio, or text: $\text{logits} = e^t \cdot (\sigma(u^{exp}) \cdot \sigma(u^{imp})^T),$
where \( t \) is a learned temperature parameter, and \( \sigma(\cdot) \) is the normalization function applied to the embeddings. The alignment loss is computed using symmetric cross-entropy, where 
$K$ represents the total number of samples in the batch:
\begin{equation}
    L_{align} = -\log \frac{e^{\text{logits}}}{\sum^K e^{\text{logits}}} - \log \frac{e^{\text{logits}^T}}{\sum^K e^{\text{logits}^T}}.
\end{equation}

During inference, the emotion embedding with the highest similarity to the implicit embeddings is selected as the final aligned emotion representation \( u^{emo} \).

\subsection{Emotion Embedding-induced TTS (EMI-TTS)}
The EMI-TTS module integrates the aligned emotion embedding \( u^{emo} \) with input text \( \text{Tex} \) to generate emotionally expressive speech using a pre-trained TTS model from the model library \( \mathcal{M} \). EMI-TTS supports a variety of TTS models, including VITS, FastSpeech2, and Tacotron2, each designed for different speech synthesis tasks.

\subsubsection{Speech Synthesis Workflow}
The process begins by encoding the input text \( \text{Tex} \) into a sequence of linguistic features \( h_{lg} \) using a text encoder:
$h_{lg} = \text{TextEncoder}(\text{Tex}).$

These linguistic features are concatenated with the emotion embedding \( u^{emo} \) and speaker embedding \( u^{spk} \) to form an augmented representation: $h_{lg}^{emo} = \text{Concat}(h_{lg}, u^{emo}, u^{spk}).$

The augmented feature is passed through the acoustic model, which generates the acoustic features \( h_{ac} \) that control prosody and other acoustic characteristics. Finally, the vocoder transforms these acoustic features into the final speech waveform: $\text{Audio}^{emo} = \text{Vocoder}(h_{ac}).$

\subsubsection{Model Variants}
The EMI-TTS framework supports several TTS model variants, each tailored for specific synthesis tasks. The following subsections detail the unique aspects of each variant, highlighting how emotion embeddings are integrated into the synthesis process.

\noindent \textbf{VITS Variant}.~The VITS variant employs a combination of variational inference and normalizing flows to incorporate emotion embeddings \( u^{emo} \) into the synthesis process. These embeddings are introduced into key components such as the Spectrogram Encoder and Emotional WaveNet (EWN) within the Emotion-condition Flow. The affine transformation operator \( \text{AX}() \) applies the emotion conditioning as follows:
\begin{equation}
\begin{aligned} 
h_0, h_1 = h &; h'_1 = \text{AX}(\text{EWN}(h_0 + u^{emo}), h_1) \\
h' &= \text{Connection}(h'_1, h_0), 
\end{aligned} 
\end{equation}
where \( h_0 \) and \( h_1 \) are hidden states, and \( h'_1 \) is the updated hidden state after emotion conditioning.

\noindent \textbf{FastSpeech2 Variant}.~In the FastSpeech2 variant, emotion embeddings are incorporated into the Conditional Cross-Attention mechanism, which merges speaker and emotion information into a unified conditioning feature \( c \). The attention mechanism uses the following formulation for queries \( Q \), keys \( K \), and values \( V \): $Q = W_q \cdot h, K = W_k \cdot c, V = W_v \cdot c,$ where \( W_q \), \( W_k \), and \( W_v \) are learnable projection matrices. The attention-weighted output is computed as:
\begin{equation}
h^{emo} = \text{softmax}\left(\frac{Q \cdot K^T}{\sqrt{d}}\right) \cdot V + h,
\end{equation}
where \( d \) is the hidden state dimension, and \( h^{emo} \) is the emotion-conditioned output. This mechanism adjusts components like the Mel-spectrogram Decoder and Duration Predictor to reflect the intended emotional tone.

\noindent \textbf{Tacotron2 Variant}.~In the Tacotron2 variant, emotion embeddings \( u^{emo} \) and speaker embeddings \( u^{spk} \) are directly integrated into the attention mechanism. The character-encoded linguistic features \( h_{lg} \) are concatenated with the emotion and speaker embeddings: $h_{lg}^{emo} = \text{Concat}(h_{lg}, u^{emo}, u^{spk}),$ and are passed through the Location \& Emotion Sensitive Attention module to generate context vectors. These vectors are then used by the Recurrent Neural Network (RNN) to produce speech that reflects both the emotional and linguistic context.

\begin{table*}[!ht]
  \caption{Comparison of WER, CER, MCD, SECS, ESMOS, SNMOS, SSMOS across TTS and audio style transfer models}
  \label{tab:WER_CER_MCD_SECS_MOS}
  \resizebox{\linewidth}{!}{
  \begin{tabular}{lcccccccccc}
    \hline 
    Model & Emotion & Dataset & WER (\%) $\downarrow$ & CER (\%) $\downarrow$ & MCD $\downarrow$ & SECS $\uparrow$  & ESMOS $\uparrow$ & SNMOS $\uparrow$ & SSMOS $\uparrow$\\
    \hline
    Ground Truth & - & ESD & 6.98 & 2.94 & / & / & $4.57(\pm 0.06)$ & $4.61(\pm 0.06)$ & $4.24(\pm 0.07)$\\
    VITS (Label) & Label & ESD & 10.82 & 5.17 & 6.285 & 0.840 & $3.80(\pm 0.09)$ & $4.09(\pm 0.08)$ & $4.09(\pm 0.09)$\\
    UMETTS (VITS) &  Prompt & ESD & 9.61 & 4.47 & \textbf{6.258} & 0.841 & $4.02(\pm 0.07)$ & \textbf{\boldmath $4.36(\pm 0.06)$} & \textbf{\boldmath $4.21(\pm 0.10)$}\\
    Emotional-TTS \cite{lee2017emotional} & Label & ESD & 34.38 & 24.30 & 10.916 & 0.760 & $2.59(\pm 0.09)$ & $2.19(\pm 0.09)$ & $2.79(\pm 0.11)$\\
    UMETTS (Tacotron) &  Prompt & ESD & 24.42 & 14.12 & 10.684 & 0.780 & $2.21(\pm 3.21)$ & $2.80(\pm 0.09)$ & $3.18(\pm 0.09)$\\
    EmoSpeech \cite{emospeech} & Label & ESD & 7.91 & 3.52 & 8.04 & 0.841 & $4.22(\pm 0.07)$ & $4.03(\pm 0.08)$ & $3.87(\pm 0.06)$\\
    GenerSpeech \cite{generspeech} &  Audio & ESD & 12.3 & 6.74 & 6.812 & 0.840 & $4.01 (\pm 0.09)$ & $3.98 (\pm 0.07)$ & $3.66 (\pm 0.08)$\\
    UMETTS (FastSpeech) &  Prompt & ESD & \textbf{7.35} & \textbf{3.07} & 6.703 & \textbf{0.896} & \textbf{\boldmath $4.37(\pm 0.07)$} & $4.29(\pm 0.06)$ & $4.13(\pm 0.07)$\\
    \hline
    Ground Truth & - & MEADTTS & 17.8 & 9.27 & / & / & $4.66 (\pm 0.09)$ & $4.57 (\pm 0.06)$ & $4.41 (\pm 0.07)$ \\
    GenerSpeech &  Audio & MEADTTS & 19.52 & 9.93 & 8.482 & 0.735 & $4.28 (\pm 0.06)$ & $3.86 (\pm 0.08)$ & $4.19 (\pm 0.09)$ \\
    MM-TTS (OOD)\cite{AAAI_MMTTS}* &  Prompt & MEADTTS & / & / & 6.69 & 0.728 & $4.25(\pm 0.22)$ & $3.62(\pm 0.43)$ & /\\
    UMETTS (ours) &  Prompt & MEADTTS & 18.7 & 9.90 & \textbf{5.927} & \textbf{0.890} & \textbf{\boldmath $4.30(\pm 0.07)$} & \textbf{\boldmath $4.08(\pm 0.07)$} & \textbf{\boldmath $4.23 (\pm 0.06)$} \\
    \hline
  \end{tabular}
  }
  MOS scores range from 1 to 5 with an interval of 0.5, with 95\% confidence intervals. SECS and MCD values represent the median. *The MEADTTS dataset is a subset of the MEAD dataset, modified as described in \cite{AAAI_MMTTS}. Due to the unavailability of the source code and complete generated samples from \cite{AAAI_MMTTS}, only demo page samples were used for evaluation.
  \vspace{-4mm}
\end{table*}

\section{Experiments}

\subsection{Experimental Setup}
\subsubsection{Datasets}
We evaluated the UMETTS framework using four widely-used emotional speech datasets: \textbf{MELD} \cite{poria2018meld}, \textbf{MEAD} \cite{kaisiyuan2020mead}, \textbf{ESD} \cite{zhou2021seen}, and \textbf{RAF-DB} \cite{li2017reliable}, which encompass text, audio, and visual modalities. \textbf{MELD} consists of 13,708 utterances (9,989 for training, 1,109 for validation, and 2,610 for testing), while \textbf{ESD} contains 17,500 utterances (14,000 for training, 1,750 for validation, and 1,750 for testing). \textbf{RAF-DB} includes 15,339 basic and 3,954 compound emotion images, split into training and testing sets. Pre-processing involved converting emotion labels into prompts. Text data were processed using \textit{Phonemizer} \cite{Bernard2021} for IPA sequences and \textit{MFA} \cite{mcauliffe2017montreal} for phoneme sequences and durations. Audio was converted into mel-spectrograms using the Short-Time Fourier Transform (STFT). Visual data, including images and video frames, were normalized and resized into tensors \(V \in \mathbb{R}^{244 \times 244 \times 3}\), with 8 frames sampled per video sequence.

\subsubsection{Model Implementation \& Training}

The vocoder used in EMI-TTS varies by model: VITS uses its original decoder, FastSpeech2 employs the \textit{iSTFTNet} vocoder \cite{kaneko2022istftnet}, and Tacotron2 utilizes a pre-trained \textit{WaveNet} \cite{van2016wavenet}, all fine-tuned on the ESD dataset. Training is divided into two phases. First, the \textit{Emotion Prompt Alignment Module (EP-Align)} is trained on multi-modal pairs from all datasets, and the resulting prompts are used to train the \textit{Emotion Embedding-induced TTS (EMI-TTS)} on audio-text pairs. Contrastive learning is used to fine-tune \textit{InstructERC} \cite{lei2023instructerc} for text, \textit{wav2vec2} \cite{baevski2020wav2vec} for audio, and \textit{ViT} \cite{dosovitskiy2020image, CLIP} for images and videos. EP-Align was trained over 100 epochs on 4 NVIDIA A100 GPUs, while EMI-TTS models were trained for 200k steps (flow-based), 40k steps (Transformer-based), and 250k steps (Recurrent-based) until convergence.

\subsubsection{Evaluation Metrics}
We assessed the UMETTS framework using both objective metrics and subjective evaluations. 

\noindent \textbf{Objective Metrics}.~We evaluated multi-modal emotion alignment accuracy, measuring how well UMETTS matches predicted emotion classes to ground-truth labels. Speech quality was assessed through several key metrics, including Word Error Rate (WER) and Character Error Rate (CER) for transcription accuracy, Mel-Cepstral Distortion (MCD) for measuring spectral differences between synthesized and reference speech, and Speaker Embedding Cosine Similarity (SECS) to evaluate how closely the synthesized voice matches the target speaker's characteristics. These metrics \footnote{For detailed implementation of these metrics, please refer to the open-source code available at https://github.com/KTTRCDL/UMETTS} were computed using the \textit{Whisper ASR} model \cite{whisper}, the \textit{mel\_cepstral\_distance} tool \cite{Sternkopf_mel-cepstral-distance_2024}, and \textit{Resemblyzer} \cite{Resemblyzer}.

\noindent \textbf{Subjective Evaluations}.~Human evaluations \footnote{we conducted subjective tests involving 10 English speakers to evaluate the performance of different models in generating synthesized audio samples. } were conducted to assess three key aspects of the synthesized speech. Emotion Similarity Mean Opinion Score (ESMOS) rated how well the synthesized speech conveyed the intended emotion. Speech Naturalness Mean Opinion Score (SNMOS) evaluated the naturalness of the speech as perceived by listeners, while Speaker Similarity Mean Opinion Score (SSMOS) measured how closely the synthesized speech matched the target speaker’s voice characteristics.

\begin{table}
\centering
  \caption{Assessing the Effect of EP-Align on Multi-modal Emotion Recognition Using Emotion Classification Accuracy}
  \label{tab:MER}
  \centering
  \begin{tabular}{ccccc}
    \hline
    Text& Audio& Images & EP-Align & F1 ($\uparrow$)\\
    \hline
    \checkmark& & &  & 0.59 \\
    & \checkmark& &  & 0.51 \\
    & & \checkmark& & 0.45 \\
    \checkmark & \checkmark & \checkmark&  & 0.68 \\
    \checkmark & \checkmark & \checkmark& \checkmark& \textbf{0.75}  \\
    \hline
  \end{tabular}
  \vspace{-5mm}
\end{table}

\subsection{Comparison with State-of-the-Art Methods}
We compared the UMETTS framework with state-of-the-art emotional TTS and audio style transfer models using both objective and subjective metrics. As shown in Table \ref{tab:WER_CER_MCD_SECS_MOS}, UMETTS consistently outperformed baseline models, especially on the ESD and MEADTTS datasets. On the \textbf{ESD dataset}, UMETTS (FastSpeech) achieved the lowest Word Error Rate (WER) of 7.35\% and Character Error Rate (CER) of 3.07\%, demonstrating superior transcription accuracy. It also recorded the highest Speaker Embedding Cosine Similarity (SECS) at 0.896, indicating strong speaker identity preservation. Subjective evaluations further highlighted UMETTS’s emotional expressiveness, with an Emotion Similarity MOS (ESMOS) of 4.37, closely matching the ground-truth score of 4.57. This demonstrates UMETTS's ability to deliver high-quality speech that preserves both emotional content and speaker identity, key for emotional TTS applications.

Similar trends were observed on the \textbf{MEADTTS dataset}. UMETTS outperformed models like GenerSpeech and MM-TTS, achieving the lowest Mel-Cepstral Distortion (MCD) of 5.927, reflecting superior spectral accuracy. In subjective evaluations, UMETTS scored higher on naturalness, confirming its ability to produce more natural and emotionally expressive speech. These results underscore UMETTS's robustness and its potential for real-world applications, where both technical precision and emotional authenticity are crucial.

\begin{figure}
    \centering
    \begin{subfigure}[b]{0.33\textwidth}
        \centering
        \includegraphics[width=\textwidth]{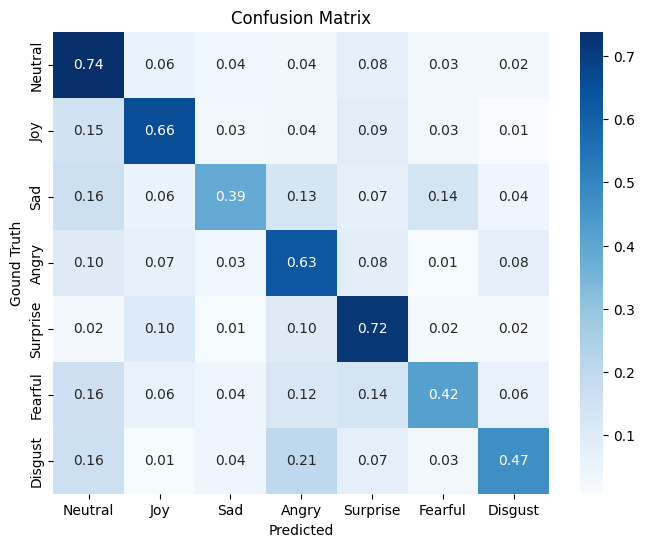}
    \end{subfigure}
    \hfill
    \begin{subfigure}[b]{0.14\textwidth}
        \centering
        \includegraphics[width=\textwidth]{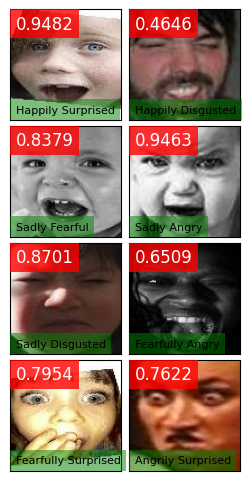}
    \end{subfigure}
    \caption{\small Left: Confusion matrix of MELD multi-modal emotion alignment with EP-Align. Right: Samples of RAF compound emotion images aligned with EP-Align.}
    \label{fig:comfusion_matrix}
    \vspace{-7mm}
\end{figure}

\subsection{Ablation Study}
\subsubsection{Impact of EP-Align on Emotion Classification}
We conducted ablation studies to evaluate the impact of EP-Align on emotion classification accuracy, using the MELD dataset for multi-modal emotion recognition and the RAF-DB dataset for compound emotion classification. The left part of Figure \ref{fig:comfusion_matrix} shows the confusion matrix for the MELD dataset, highlighting EP-Align’s performance in predicting basic and compound emotions. Misclassifications, such as the overlap between 'joy' and 'surprise', reveal areas for improvement. On the right side of Figure \ref{fig:comfusion_matrix}, we show cosine similarity between image embeddings and prompt embeddings for randomly selected images from RAF with original labels. These insights provide guidance for refining EP-Align to improve sensitivity to subtle emotional differences.

\subsubsection{Effect of Emotion Labels vs. Aligned Prompts on Speech Quality}
We further examined the difference in speech quality generated by various models using Emotion Labels versus Aligned Prompts. As shown in Table \ref{fig:comfusion_matrix}, EP-Align effectively leverages emotional cues from multiple modalities, enabling the generation of highly expressive and emotionally resonant speech. By incorporating superior emotional representations, EP-Align enhances both the intelligibility and naturalness of synthesized speech across different models, showcasing its effectiveness in multi-modal emotion integration.

\section{conclusions and discussions}

In this work, we introduce UMETTS, a novel multimodal emotional text-to-speech synthesis framework that effectively integrates emotional cues from multiple modalities. Our evaluations demonstrated that UMETTS consistently outperforms traditional models in both objective and subjective metrics, highlighting its potential for applications requiring emotionally rich speech synthesis. By open-sourcing our work, we hope to enable further advances in the field and contribute to the development of emotionally intelligent human-computer interactions.


\bibliographystyle{IEEEtran}
\bibliography{reference}

\end{document}